\documentclass{article}
  
\usepackage{multirow}
\usepackage[preprint]{nips_2018}
 
\usepackage[utf8]{inputenc} 
\usepackage[T1]{fontenc}    
\usepackage{hyperref}       
\usepackage{url}            
\usepackage{booktabs}       
\usepackage{amsfonts}       
\usepackage{nicefrac}       
\usepackage{microtype}      
\usepackage{graphicx}
\usepackage{tabularx}
\usepackage{amssymb,amsmath}
\usepackage{float}

\newcommand\setrow[1]{\gdef\rowmac{#1}#1\ignorespaces}
\newcommand\clearrow{\global\let\rowmac\relax}
\clearrow

\title{Multi-function Convolutional Neural Networks for Improving Image Classification Performance}

\begin{document}
\author {Luna M. Zhang}
\maketitle

\begin{abstract}
Traditional Convolutional Neural Networks (CNNs) typically use the same activation function (usually ReLU) for all neurons with non-linear mapping operations. For example, the deep convolutional architecture Inception-v4 uses ReLU. To improve the classification performance of traditional CNNs, a new "Multi-function Convolutional Neural Network" (MCNN) is created by using different activation functions for different neurons. For $n$ neurons and $m$ different activation functions, there are a total of $m^n-m$ MCNNs and only $m$ traditional CNNs. Therefore, the best model is very likely to be chosen from MCNNs because there are $m^n-2m$ more MCNNs than traditional CNNs. For performance analysis, two different datasets for two applications (classifying handwritten digits from the MNIST database and classifying brain MRI images into one of the four stages of Alzheimer's disease (AD)) are used. For both applications, an activation function is randomly selected for each layer of a MCNN. For the AD diagnosis application, MCNNs using a newly created multi-function Inception-v4 architecture are constructed. Overall, simulations show that MCNNs can outperform traditional CNNs in terms of multi-class classification accuracy for both applications. An important future research work will be to efficiently select the best MCNN from $m^n-m$ candidate MCNNs. Current CNN software only provides users with partial functionality of MCNNs since different layers can use different activation functions but not individual neurons in the same layer. Thus, modifying current CNN software systems such as ResNets, DenseNets, and Dual Path Networks by using multiple activation functions and developing more effective and faster MCNN software systems and tools would be very useful to solve difficult practical image classification problems.  
\end{abstract}

\section{Introduction}

Deep learning techniques are very effective for computer vision applications [1-6]. In particular, in recent years, Convolutional Neural Networks (CNNs) have been successful for image classification in various important real-world applications such as medical imaging [4], self-driving cars [5], and board games [6]. For example, CNNs detected skin cancer (about 72.1\% accuracy) better than dermatologists did (about 66.0\% accuracy) in 2017 [4]. 

In 1979, Fukushima developed the earliest CNN [7]. It had several convolutional and pooling layers similar to modern CNNs. In 1998, LeCun et al. developed LeNet, which had outstanding performance for handwritten digit recognition and zip codes [8]. AlexNet had an architecture that was very similar to LeNet, but it was larger and deeper, and had convolutional layers positioned on top of each other [2]. In 2012, AlexNet was used for the ImageNet ILSVRC competition to significantly outperform the second runner-up [9]. 

In recent years, many powerful popular CNNs have been created with outstanding performance results that include results from important competitions such as the ImageNet Large Scale Visual Recognition Competition (ILSVRC) [9-12] and COCO Object Detection Task Competition [13-14]. Some examples are GoogLeNet [15], ResNets [3], DenseNets [16], Dual Path Networks (DPNs) [17], and Inception-v4 networks [18]. Each of these specific CNNs (and not just limited to these CNNs) uses the same activation function (ReLU) for all of the non-linear operations after convolution. However, using different activation functions in the same CNN may achieve better performance than using ReLU.
 
Traditional CNNs use the same activation function for the activation operations followed by convolutional layers and for the fully-connected layers (FCLs). ReLU is a popular choice since it has many good properties, such as being simple and having a non-vanishing gradient. However, there are other activation functions that may be a better choice than ReLU. For example, the "exponential linear unit" (ELU) is an activation function that has been shown to speed up learning and outperform ReLU networks in terms of classification accuracy [19].  A neural network using different activation functions can achieve better training and testing performance than a neural network using one activation function [20]. Not only can CNNs using an activation function other than ReLU perform better, but also CNNs can use a variety of different activation functions. Thus, we propose a new "Multi-function CNN" (MCNN) that uses different activation functions for neurons to improve performance such as classification accuracy. It will be important to create new MCNN software that can automatically and quickly find an optimal or near-optimal set of activation functions for different neurons to effectively improve image classification performance.

\section{Multi-function Convolutional Neural Networks and Single-function Convolutional Neural Networks}

\subsection{Convolutional Layers}

The purpose of convolution is to extract features from the input image. A filter slides over the input image (convolution operation) to produce an output image that is commonly called a feature map of convolved features. There can be many convolution layers in a CNN.
 
Suppose we have a $W\times H \times D$ image, and we apply $m$ filters ($F_1, F_2, ..., F_m$) of dimension $N\times N\times D$ that are convolved with the input image. Let a filter have $D$ kernels: $D$ $N\times N$ matrices $w_{k}$ where  $k=1, 2, ..., D$. Let $x$ be a $N\times N\times D$ patch. Let $b^n$ be the bias of a filter $F_n$. 

A convolved feature of a feature map among $m$ feature maps after applying a filter $F_n$ for $n = 1, 2, ..., m$ is:
$$\theta^n = \sum_{k=1}^{D}\sum_{i=1}^{N}\sum_{j=1}^{N} w^n_{ijk}x_{ijk}+b^n.$$

Similarly, the rest of the convolved features are calculated using the above formula.

\subsection{Activation Layers}

Seven typical activation functions (with their 3-letter abbreviations for later reference) are Linear (LIN), Rectifier (REL), Sigmoid (SIG), Hyperbolic Tangent (TAN), Softplus (PLS), Softsign (SGN), and Exponential Linear Unit (ELU).

A typical CNN uses a ReLU layer right after each convolutional layer (CONV) and before a pooling layer (POOL) to generate $m$ new non-linear feature maps from the $m$ feature maps. For example, a typical CNN architecture is: INPUT $\rightarrow$ [[CONV $\rightarrow$ REL] * M $\rightarrow$ POOL] * N $\rightarrow$ FCL $\rightarrow$ OUTPUT. Let * indicate repetition and M $\geqslant$ 1 and N $\geqslant$ 1. The same activation function (REL) is used to map all convolved features to new features. However, different activation functions may be used to achieve better classification accuracy.

An activation layer (AL) is defined as a layer of neurons where each neuron uses an activation function selected from a set of different activation functions. An AL transforms $m$ feature maps to new $m$ feature maps. The convolved feature $\theta^n$ is transformed to a new feature, which is $f(\theta^n)$, by an activation function $f$. Different activation functions can be used for different individual convolved features and different feature sub-maps.

For example, let AL(REL, SIG) mean that in the same AL, some neurons use REL and the others use SIG. A new MCNN uses an AL right after each CONV and before POOL.

Let a convolution block (CB) be a CONV followed by an AL. For example, a MCNN's CB is [CONV $\rightarrow$ AL(REL, SIG)]. The traditional CNN's CB is [CONV $\rightarrow$ REL], which is equivalent to the newly defined notation: [CONV $\rightarrow$ AL(REL)]. AL(REL) means that all neurons on the AL use REL.

An example of a MCNN architecture is INPUT $\rightarrow$ [CONV $\rightarrow$ AL(REL, SIG)] $\rightarrow$ POOL $\rightarrow$ [CONV $\rightarrow$ AL(REL, SIG, TAN)] $\rightarrow$ [CONV $\rightarrow$ AL(LIN)] $\rightarrow$ POOL $\rightarrow$ [CONV $\rightarrow$ AL(PLS, SIG)] $\rightarrow$ [CONV $\rightarrow$ AL(SGN, ELU)] $\rightarrow$ [CONV $\rightarrow$ AL(TAN, ELU, LIN)] $\rightarrow$ POOL $\rightarrow$ FCL $\rightarrow$ OUTPUT. Another example for a MCNN architecture is INPUT $\rightarrow$ [CONV $\rightarrow$ AL(REL)] $\rightarrow$ POOL $\rightarrow$ [CONV $\rightarrow$ AL(SIG)] $\rightarrow$ [CONV $\rightarrow$ AL(TAN)]  $\rightarrow$ POOL $\rightarrow$ FCL $\rightarrow$ OUTPUT. 

\subsection{Multi-function Inception-v4}
 
Inception-v4 is an effective CNN architecture. For example, an ensemble system of three ResNets and one Inception-v4 network achieved 3.08\% top-5
error on the test set of the ImageNet classification challenge [18]. A 
CB using REL is a basic building block for constructing an Inception-v4 network. For example, Fig. 1 shows that every CB of an Inception-A architecture uses the same activation function (REL). 
 
\begin{figure}[H] 

  \centerline{\includegraphics[width=10cm, height=7cm, keepaspectratio]{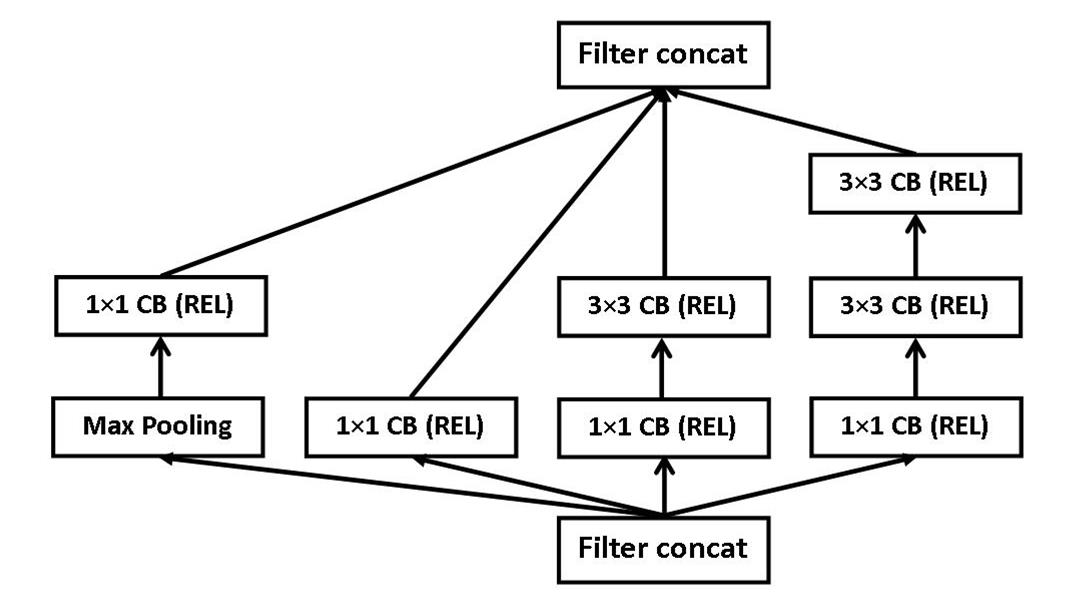}
}
  \caption{An Inception-A architecture based on Inception-v4 [18].}
  \label{fig: Inception-A }
\end{figure}

The newly created multi-function Inception-v4 consists of input, multi-function stem, 4$\times$ multi-function Inception-A, multi-function Reduction-A, 7$\times$ multi-function Inception-B, multi-function Reduction-B, 3$\times$ multi-function Inception-C, Max Pooling, Dropout(0.8), and Softmax. It does not have FCLs. For example, Fig. 2 shows that different CBs of a multi-function Inception-A architecture use 7 different activation functions.  

\begin{figure}[H] 

  \centerline{\includegraphics[width=10cm, height=7cm, keepaspectratio]{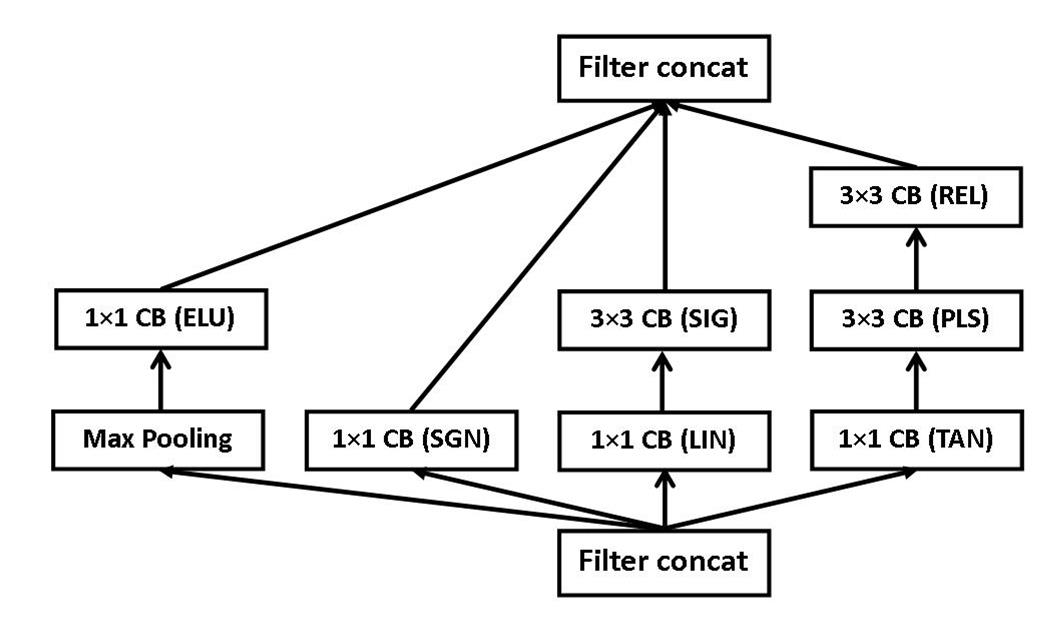}
}
  \caption{A multi-function Inception-A architecture based on Inception-v4 [18].}
  \label{fig: Inception-A}
\end{figure} 

Let a convolutional path be a series of connected max pooling and/or CBs starting from a filter concatenation stage to the next filter concatenation stage. A simple multi-function Inception-A is shown in Fig. 3, which has four convolutional paths (Path 1: Max Pooling $\rightarrow$ 1$\times$1 CB(REL), Path 2: 1$\times$1 CB(SIG), Path 3: 1$\times$1 CB(TAN) $\rightarrow$ 3$\times$3 CB(TAN), and Path 4: 1$\times$1 CB(REL) $\rightarrow$ 3$\times$3 CB(REL) $\rightarrow$ 3$\times$3 CB(REL)). An AL in the CB on Path 1 uses REL, an AL in the CB on Path 2 uses SIG, two ALs in the two CBs on Path 3 use TAN, and three ALs in the three CBs on Path 4 use REL. A simple method for building a multi-function Inception-v4 architecture randomly assigns an activation function to all ALs in all CBs on each convolutional path. 
\begin{figure}[H] 

  \centerline{\includegraphics[width=10cm, height=7cm, keepaspectratio]{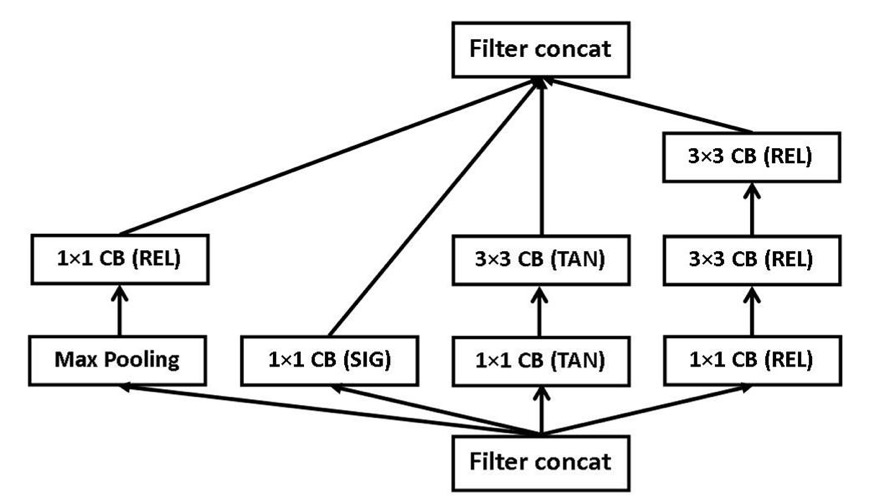}
}
  \caption{A simple multi-function Inception-A architecture based on Inception-v4 [18].}
  \label{fig: Inception-A }
\end{figure}

\subsection{Different types of MCNN and SCNN models}

Let there be a total of $m$ AL neurons (each neuron uses an activation function selected from a set of $j$ activation functions) and $n$ fully-connected (FC) hidden neurons (each neuron uses an activation function selected from a set of $k$ activation functions). For example, three FC hidden neurons in the first FC hidden layer may use SIG, LIN, and REL, and a FC hidden neuron in the third FC hidden layer may use TAN. Let a traditional CNN, which uses a single activation function for all neurons, be called a "Single-function CNN" (SCNN). Let the four types of general CNNs be denoted as SCNN-SS, MCNN-SM, MCNN-MS, and MCNN-MM, and for each type, the total numbers of CNNs based on different function combinations are given in Table 1. Let "multi-function" mean that at least two different activation functions are used for different neurons. Let "single-function" mean that only one activation function is used for all neurons. Note that all neurons in a SCNN-SS should not all use LIN since the entire transformation from the input to the final output should be non-linear.

\begin{table}[!htb]
  \caption{Four types of general CNNs}
  \label{table1}
  \centering
  \begin{tabular}{>{\rowmac}c>{\rowmac}c>{\rowmac}c>{\rowmac}c<{\clearrow}}
    \toprule
    CNN Type & AL Neurons & FC Hidden Neurons & No. CNNs\\
 \midrule
   SCNN-SS & single-function & single-function & \textit{jk}   \\
     MCNN-SM & single-function & multi-function & \textit{j($k\textsuperscript{n}$-k)}\\
     MCNN-MS & multi-function  & single-function  & \textit{($j\textsuperscript{m}$-j)k} \\
     MCNN-MM & multi-function  & multi-function & \textit{($j\textsuperscript{m}$-j)($k\textsuperscript{n}$-k)}   \\
    \bottomrule
  \end{tabular}
\end{table}

Table 2 shows the case for CNNs without FCLs. Let the two types of general CNNs without FCLs be denoted as SCNN-S and MCNN-M.

\begin{table}[!htb]
  \caption{Two types of general CNNs without FCLs}
  \label{table2}
  \centering
  \begin{tabular}{>{\rowmac}c>{\rowmac}c>{\rowmac}c<{\clearrow}}
    \toprule
    CNN Type & AL Neurons & No. CNNs\\
 \midrule
   SCNN-S & single-function  & \textit{j}   \\
     MCNN-M & multi-function  & \textit{$j\textsuperscript{m}$-j}\\
    \bottomrule
  \end{tabular}
\end{table}

\subsection{New general algorithms for creating MCNNs}

The following four general algorithms are developed to create the four types of MCNNs: MCNN-SM, MCNN-MS, MCNN-MM, and MCNN-M.

Algorithm 1 - Creating a MCNN-SM:
  1) Build CBs where all AL neurons use the same activation function.
  2) Build FC hidden layers where FC hidden neurons use different activation functions.
  3) Build a complete MCNN-SM using the built CBs and FC hidden layers.

Algorithm 2 - Creating a MCNN-MS:
  1) Build CBs where all AL neurons use different activation functions.
  2) Build FC hidden layers where FC hidden neurons use the same activation function.
  3) Build a complete MCNN-MS using the built CBs and FC hidden layers.

Algorithm 3 - Creating a MCNN-MM:
  1) Build CBs where all AL neurons use different activation functions.
  2) Build FC hidden layers where FC hidden neurons use different activation functions.
  3) Build a complete MCNN-MM using the built CBs and FC hidden layers.
  
Algorithm 4 - Creating a MCNN-M:
  1) Build CBs where all AL neurons use different activation functions.
  2) Build a complete MCNN-M using the built CBs.

\section{Experimental results}

SCNNs and MCNNs with both ALs and FCLs are created and tested for the first application (classifying ten handwritten digits from the MNIST database). SCNNs and MCNNs with only ALs are also created and tested for the second application (classifying brain MRI images into one of the four stages of Alzheimer's disease (AD)). Python scripts were written using Keras and scikit-learn. 

\subsection{Application 1: classifying ten handwritten digits from the MNIST database}

The MNIST dataset is a well-known classic benchmark dataset for classification algorithms in computer vision [21-22]. There are 42,000 training images and 28,000 testing images. 3 to 8 FC hidden layers were used, and 32, 64, or 128 neurons were used in each FC hidden layer. Six activation functions were used for ALs and FC hidden layers: LIN, REL, SIG, TAN, PLS, and SGN. For all CNNs tested, the output layer used the softmax function. Current CNN software limitations include an activation function being unable to be chosen for a specific neuron, so all neurons in the same AL or FC hidden layer must use the same activation function. SCNN-SS, MCNN-SM, MCNN-MS, and MCNN-MM models were compared in terms of test accuracy for the 10-class classification. For comparison, all MCNN models are bolded, and the models are ordered by decreasing test accuracy in Tables 3-8. For example, for any table, the first row represents the CNN with the highest test accuracy and the last row represents the CNN with the lowest test accuracy. Let ``HL1'' denote the first FC hidden layer, ``HL2'' denote the second FC hidden layer, etc. Let ``AL1'' denote the first activation layer, ``AL2'' denote the second activation layer, etc.

\subsubsection{Simulation results for MCNN-SM and SCNN-SS models}

Tables 3-6 show results for MCNN-SM and SCNN-SS models, for 5 to 8 FC hidden layers and 3 ALs (AL1 - 64 neurons, AL2 - 128 neurons, AL3 - 128 neurons) using REL. For example, in Table 3, for the first MCNN-SM model (REL-LIN-REL-LIN-REL), all neurons in HL1 use REL, all neurons in HL2 use LIN, all neurons in HL3 use REL, all neurons in HL4 use LIN, and all neurons in HL5 use REL. 

\begin{table}[h!]
  \caption{Performance of MCNN-SM and SCNN-SS models with 5 FC hidden layers}
  \label{table3}
  \centering
  \begin{tabular}{>{\rowmac}c>{\rowmac}c>{\rowmac}c>{\rowmac}c>{\rowmac}c>{\rowmac}c>{\rowmac}c<{\clearrow}}
    \toprule
    HL1& HL2& HL3& HL4& HL5 & Training Accuracy & Test Accuracy\\
 \midrule
   \setrow{\bfseries}REL& LIN& REL& LIN& REL & 99.940\%& 99.443\%  \\
  REL& REL& REL& REL& REL & 99.940\%& 99.386\%   \\
    \setrow{\bfseries}REL& PLS& SIG& TAN& REL & 99.910\%& 99.371\%\\
    LIN& LIN& LIN& LIN& LIN & 99.940\%& 99.357\% \\
    \setrow{\bfseries}REL& REL& LIN& LIN& LIN & 99.910\%& 99.329\% \\
     \setrow{\bfseries}REL& LIN& REL& LIN& LIN & 99.950\%& 99.300\% \\
PLS& PLS& PLS& PLS& PLS & 99.940\%& 99.229\%\\    
    TAN& TAN& TAN& TAN& TAN & 99.750\%& 99.157\% \\
     \setrow{\bfseries}REL& SIG& SOF& PLS& TAN & 99.940\%& 99.157\%\\
     SIG& SIG& SIG& SIG& SIG & 99.940\%& 99.114\% \\
    \bottomrule
  \end{tabular}
\end{table}

\begin{table}[h!]
  \caption{Performance of MCNN-SM and SCNN-SS models with 6 FC hidden layers}
  \label{table4}
  \centering
  \begin{tabular}{>{\rowmac}c>{\rowmac}c>{\rowmac}c>{\rowmac}c>{\rowmac}c>{\rowmac}c>{\rowmac}c>{\rowmac}c<{\clearrow}}
    \toprule
    HL1&  HL2& HL3& HL4& HL5& HL6 & Training Accuracy & Test Accuracy\\
 \midrule
   \setrow{\bfseries}REL& LIN& REL& LIN& REL& PLS & 99.960\%& 99.543\%  \\
  \setrow{\bfseries}REL& LIN& REL& LIN& REL& LIN & 99.940\%& 99.443\%   \\
    \setrow{\bfseries}PLS& LIN& REL& LIN& REL& PLS & 99.900\%& 99.414\%  \\
    \setrow{\bfseries}REL& LIN& PLS& LIN& PLS& PLS & 99.970\%& 99.400\% \\
    PLS& PLS& PLS& PLS& PLS& PLS & 99.910\%& 99.386\% \\
    REL& REL& REL& REL& REL& REL & 99.950\%& 99.371\% \\
    LIN& LIN& LIN& LIN& LIN& LIN & 99.950\%& 99.186\% \\
    \bottomrule
  \end{tabular}
\end{table}

\begin{table}[h!]
  \caption{Performance of MCNN-SM and SCNN-SS models with 7 FC hidden layers}
  \label{table5}
  \centering
  \begin{tabular}{>{\rowmac}c>{\rowmac}c>{\rowmac}c>{\rowmac}c>{\rowmac}c>{\rowmac}c>{\rowmac}c>{\rowmac}c>{\rowmac}c<{\clearrow}}
    \toprule
    HL1& HL2& HL3& HL4& HL5& HL6& HL7 & Training Accuracy & Test Accuracy\\
 \midrule
   \setrow{\bfseries}LIN& LIN& PLS& LIN& LIN& REL& REL & 99.962\%& 99.400\%\\
\setrow{\bfseries}LIN& LIN&LIN& PLS& REL& REL& REL& 99.971\%& 99.314\% \\
REL& REL& REL& REL& REL& REL& REL & 99.971\%& 99.286\%\\
\setrow{\bfseries}LIN& LIN& PLS& PLS& PLS& REL& REL & 99.952\%& 99.286\%\\
\setrow{\bfseries}REL& LIN& REL& LIN& REL& LIN& REL & 99.964\%& 99.257\%\\
LIN& LIN& LIN& LIN& LIN& LIN& LIN & 99.969\%,& 99.143\%\\
PLS& PLS& PLS& PLS& PLS& PLS& PLS & 99.950\%& 98.986\%\\
    \bottomrule
  \end{tabular}
\end{table}

\begin{table}[h!]
  \caption{Performance of MCNN-SM and SCNN-SS models with 8 FC hidden layers}
  \label{table6}
  \centering
  \begin{tabular}{>{\rowmac}c>{\rowmac}c>{\rowmac}c>{\rowmac}c>{\rowmac}c>{\rowmac}c>{\rowmac}c>{\rowmac}c>{\rowmac}c>{\rowmac}c<{\clearrow}}
    \toprule
    HL1& HL2& HL3& HL4& HL5& HL6& HL7& HL8 & Training Acc. &Test Acc.\\
 \midrule
   \setrow{\bfseries}REL& REL& LIN& REL& REL& REL& LIN& REL & 99.920\%& 99.500\% \\
    \setrow{\bfseries}REL& LIN& PLS& LIN& REL& REL& PLS& REL & 99.920\%& 99.457\% \\
   \setrow{\bfseries}REL& PLS& LIN&REL& REL& REL& REL& REL & 99.930\%& 99.429\% \\
   \setrow{\bfseries}REL& LIN& PLS& LIN& PLS& LIN& REL& REL & 99.990\%& 99.400\% \\
   \setrow{\bfseries}REL& REL& LIN& REL& LIN& REL& LIN& REL & 99.980\%& 99.371\% \\
     REL& REL& REL& REL& REL& REL& REL& REL & 99.962\%& 99.357\% \\
     PLS& PLS& PLS& PLS& PLS& PLS& PLS& PLS & 99.962\%& 99.243\%  \\
     LIN& LIN& LIN& LIN& LIN& LIN& LIN& LIN & 99.943\%& 99.186\%  \\
    \bottomrule
  \end{tabular}
\end{table}

\subsubsection{Simulation results for MCNN-MS and SCNN-SS models}

Table 7 shows results for MCNN-MS and SCNN-SS models, for 3 ALs and 4 FC hidden layers (no. neurons: AL1-64, AL2-128, AL3-128, HL1-64, HL2-64, HL3-32, HL4-64).\\

\begin{table}[h!]
  \caption{Performance of MCNN-MS and SCNN-SS models with 3 ALs and 4 FC hidden layers}
  \label{table7}
  \centering
  \begin{tabular}{>{\rowmac}c>{\rowmac}c>{\rowmac}c>{\rowmac}c>{\rowmac}c>{\rowmac}c<{\clearrow}}
    \toprule
    AL1& AL2& AL3 & HL1-HL4 & Training Accuracy & Test Accuracy\\
 \midrule
   \setrow{\bfseries}REL& SGN& REL & SGN & 99.950\%& 99.400\% \\
   \setrow{\bfseries}REL& TAN& LIN & PLS & 99.957\%& 99.357\% \\
   \setrow{\bfseries}REL& SGN& REL & REL & 99.955\%& 99.342\% \\
   \setrow{\bfseries}REL& LIN& REL & LIN & 99.967\%& 99.328\% \\
   REL& REL& REL & REL & 99.950\%& 99.300\% \\
   REL& REL& REL & SIG & 99.955\%& 99.286\% \\
   REL& REL& REL & TAN & 99.957\%& 99.286\% \\
   \setrow{\bfseries}LIN& LIN& REL & REL & 99.976\%& 99.285\%\\
    \bottomrule
  \end{tabular}
\end{table}

\subsubsection{Simulation results for MCNN-MM and SCNN-SS models}

Table 8 shows results for MCNN-MM and SCNN-SS models, for 2 ALs and 4 FC hidden layers (no. neurons: AL1-64, AL2-128, HL1-64, HL2-64, HL3-32, HL4-64).\\

\begin{table}[h!]
  \caption{Performance of MCNN-MM and SCNN-SS models with 2 ALs and 4 FC hidden layers}
  \label{table8}
  \centering
  \begin{tabular}{>{\rowmac}c>{\rowmac}c>{\rowmac}c>{\rowmac}c>{\rowmac}c>{\rowmac}c>{\rowmac}c>{\rowmac}c<{\clearrow}}
    \toprule
    AL1& AL2 & HL1& HL2& HL3& HL4 & Training Accuracy & Test Accuracy\\
 \midrule
   \setrow{\bfseries}LIN& REL & PLS& REL& LIN& PLS & 99.302\%& 98.957\% \\
   \setrow{\bfseries}LIN& REL & LIN& SIG& REL& PLS & 99.414\%& 98.928\% \\
  	REL& REL & PLS& PLS& PLS& PLS & 99.429\%& 98.871\% \\
  	\setrow{\bfseries}LIN& REL &LIN& SIG& SGN& REL & 99.474\%& 98.857\% \\
	REL& REL & REL& REL& REL& REL & 99.317\%& 98.842\% \\
	REL& REL & LIN& LIN& LIN& LIN & 99.510\%& 98.828\% \\    
\setrow{\bfseries}LIN& REL & LIN& SIG& REL& SGN & 99.317\%& 98.814\% \\  
    \bottomrule
  \end{tabular}
\end{table}

\subsubsection{Performance analysis}

For the simulations, a small number of MCNN models were trained, and there was an average of 3 MCNN models better than the best SCNN models based on Tables 3-8. Therefore, it is relatively easy to find a MCNN model with better performance than a SCNN model, so it would be feasible to perform MCNN model selection to quickly identify the best MCNN model.

\paragraph{Comparison between MCNN-SM and SCNN-SS}

Overall, there were 2 (out of 6) best-ranked MCNN-SM models that got higher training accuracies than the best-ranked SCNN-SS models. SCNN-SS using REL is ranked 4.25 on average based on Tables 3-6. For one case shown in Table 4, SCNN-SS using PLS is better than SCNN-SS using REL in terms of test accuracy. Thus, a CNN using a popular REL is not always optimal. For both training accuracies and test accuracies in Tables 3-6, the best-ranked MCNN-SM models consistently placed 1st whereas the best-ranked SCNN-SS models did not. Overall, all 6 best-ranked MCNN-SM models got higher test accuracies than the best-ranked SCNN-SS models. There are test accuracy improvements of the best-ranked MCNN-SM model over the best-ranked SCNN-SS model for every table. The differences in the test accuracies range from approximately 0.01\% to 0.16\%.

\paragraph{Comparison between MCNN-MS and SCNN-SS}

For testing, the best-ranked MCNN-MS model is more accurate than the best-ranked SCNN-SS model by 0.1\%. The top 4 are MCNN models, and a SCNN-SS model using REL is ranked 5th as shown in Table 7.

\paragraph{Comparison between MCNN-MM and SCNN-SS}

For testing, the best-ranked MCNN-MM model is more accurate than the best-ranked SCNN-SS model by 0.086\%. The top 4 are MCNN models, and a SCNN-SS model using REL is ranked 5th as shown in Table 8.

\subsection{Application 2: Diagnosing Alzheimer's disease using brain MRI images}

An AD dataset with 436 MRI brain images (with extra data for 20 subjects) [23], which is pre-processed and ready to be used, is used for performance analysis. This dataset has a cross-sectional collection of 416 subjects aged 18 to 96. This research work uses all brain MRI images for a 4-class classification problem to determine the AD stage (non-demented, very mild dementia, mild dementia, or moderate dementia) of a person [23-24].

Stratified 3-fold cross validation was used to evaluate and compare SCNNs and MCNNs with the calculations of comprehensive multi-class classification metrics (i.e. training accuracies, test accuracies, train F1-scores and test F1-scores). An activation function set \{REL, SIG, TAN\} was used to build different MCNN-M and SCNN-S models.

Inception-v4 was used to build a SCNN-S model using REL. Inception-v4 was modified to use SIG or TAN instead of REL to create a SCNN-S model using SIG or TAN. MCNN-M models were built by using the multi-function Inception-v4 using convolutional paths as shown in Fig. 3. 

\subsubsection{Simulation results}

In Table 9, the top-ranked MCNN-M model with the multi-function Inception-v4 (MCNN-M model \#1) has 65 convolutional paths (1, 1, 3, 2, 1, 1, 1, 2, 1, 3, 1, 2, 1, 1, 1, 3, 3, 3, 1, 1, 1, 2, 1, 3, 3, 2, 3, 2, 1, 3, 2, 2, 3, 3, 3, 1, 1, 3, 1, 3, 2, 1, 2, 1, 3, 3, 3, 1, 3, 2, 3, 3, 1, 3, 1, 1, 2, 1, 2, 3, 2, 2, 2, 2, 2 where 1 means that all ALs in all CBs on a convolutional path use REL, 2 means that all ALs in all CBs on a convolutional path use SIG and 3 means that all ALs in all CBs on a convolutional path use TAN). A SCNN-S model with Inception-v4 uses only one activation function for all ALs in all CBs on every convolutional path. In Table 9, the top-ranked SCNN-S model \#1 uses REL, SCNN-S model \#2 uses SIG, and SCNN-S model \#3 uses TAN.

\begin{table}[h!]
  \caption{Top MCNN-M Models and Top SCNN-S Models with 7 training epochs}
  \label{table12}
  \centering
  \begin{tabular}{ccccccccc}
    \toprule
    &\multicolumn{4}{c}{MCNN-M Model}
&&
\multicolumn{3}{c}{SCNN-S Model} \\\cmidrule(r){2-5}\cmidrule(l){6-9}
     & \#1 & \#2 & \#3 & \#4 && \#1(REL) & \#2(SIG) & \#3(TAN)\\
 \midrule
   Training Acc. & 79.68\% & \textbf{80.16\%} & 79.51\% & 74.07\% && 78.85\% & 74.39\% & 42.77\%\\
   Test Accuracy & \textbf{79.68\%} & 78.68\% & 77.41\% & 70.87\% && 78.03\% & 75.48\% & 39.36\%\\
   Train F1-Score & 0.7066 & 0.7228 & \textbf{0.7337} & 0.7057 && 0.7079 & 0.6909 & 0.4968\\
   Test F1-Score & 0.7068 & 0.7091 & \textbf{0.7156} & 0.6821 && 0.7054 & 0.6999 & 0.4843\\
    \bottomrule
  \end{tabular}
\end{table}

In Table 10, for MCNN-M models, REL, SIG, and TAN are used. SCNN-S model \#1 uses SIG, SCNN-S model \#2 uses REL, and SCNN-S model \#3 uses TAN. 

\begin{table}[h!]
  \caption{Top MCNN-M Models and Top SCNN-S Models with 8 training epochs}
  \label{table13}
  \centering
  \begin{tabular}{ccccccccc}
    \toprule
    &\multicolumn{4}{c}{MCNN-M Model}
&&
\multicolumn{3}{c}{SCNN-S Model} \\\cmidrule(r){2-5}\cmidrule(l){6-9}
     & \#1 & \#2 & \#3 & \#4 &&  \#1(SIG) & \#2(REL) & \#3(TAN)\\
 \midrule
   Training Acc. &  84.92\% & 84.92\% & \textbf{85.25\%} & 83.45\% && 84.92\% & 85.08\% & 42.62\%\\
   Test Accuracy & \textbf{84.93\%} & \textbf{84.93\%} & 84.27\% & 82.61\% && \textbf{84.93\%} & 84.59\% & 37.40\%\\
   Train F1-Score &  0.7800 & 0.7800 & \textbf{0.8003} & 0.7889 && 0.7800 & 0.7837 & 0.5167\\
   Test F1-Score & 0.7801 & 0.7801 & \textbf{0.7861} & 0.7698 && 0.7801 & 0.7784 & 0.4797\\
    \bottomrule
  \end{tabular}
\end{table}

In Table 11, for MCNN-M models, REL, SIG, and TAN are used. SCNN-S model \#1 uses REL, SCNN-S model \#2 uses SIG, and SCNN-S model \#3 uses TAN. 

\begin{table}[h!]
  \caption{Top MCNN-M Models and Top SCNN-S Models with 9 training epochs}
  \label{table13}
  \centering
  \begin{tabular}{ccccccccc}
    \toprule
    &\multicolumn{4}{c}{MCNN-M Model}
&&
\multicolumn{3}{c}{SCNN-S Model} \\\cmidrule(r){2-5}\cmidrule(l){6-9}
     & \#1 & \#2 & \#3 & \#4 &&  \#1(REL) & \#2(SIG) & \#3(TAN)\\
 \midrule
   Training Acc. &  72.97\% & 69.71\% & 67.40\% & 61.19\% && \textbf{73.74\%} & 61.62\% & 56.80\%\\
   Test Accuracy & \textbf{73.74\%} & 64.84\% & 63.23\% & 62.22\% && 72.84\% & 54.29\% & 50.53\%\\
   Train F1-Score &  0.7037 & \textbf{0.7164} & 0.6934 & 0.6318 && 0.7145 & 0.5816 & 0.6664\\
   Test F1-Score & \textbf{0.7076} & 0.6735 & 0.6628 & 0.6338 && 0.7070 & 0.5561 & 0.5664\\
    \bottomrule
  \end{tabular}
\end{table}

In Table 12, for MCNN-M models, REL, SIG, and TAN are used. SCNN-S model \#1 uses REL, SCNN-S model \#2 uses SIG, and SCNN-S model \#3 uses TAN. 

\begin{table}[h!]
  \caption{Top MCNN-M Models and Top SCNN-S Models with 10 training epochs}
  \label{table13}
  \centering
  \begin{tabular}{ccccccccc}
    \toprule
    &\multicolumn{4}{c}{MCNN-M Model}
&&
\multicolumn{3}{c}{SCNN-S Model} \\\cmidrule(r){2-5}\cmidrule(l){6-9}
     & \#1 & \#2 & \#3 & \#4 &&  \#1(REL) & \#2(SIG) & \#3(TAN)\\
 \midrule
   Training Acc. &  \textbf{81.31\%} & 79.17\% & 79.34\% & 76.42\% && 77.71\% & 64.48\% & 54.89\%\\
   Test Accuracy & \textbf{79.34\%} & 75.75\% & 75.74\% & 72.44\% && 76.38\% & 61.54\% & 46.21\%\\
   Train F1-Score &  \textbf{0.7757} & 0.7699 & 0.7734 & 0.7741 && 0.7095 & 0.6490 & 0.6174\\
   Test F1-Score & \textbf{0.7557} & 0.7297 & 0.7377 & 0.7322 && 0.6158 & 0.6158 & 0.5552\\
    \bottomrule
  \end{tabular}
\end{table}

\subsubsection{Performance analysis}

From Tables 9-12, MCNN-M models can outperform SCNN-S models since the best MCNN-M models achieve better test accuracies than the best SCNN-S models for three cases, the two best MCNN-M models have the same test accuracy as the best SCNN-S model for one case, the best MCNN-M model has higher train F1-scores and test F1-scores than the best SCNN-S model for all four cases. Best MCNN-M models are better than SCNN-S models using REL based on Tables 9-12. For one case shown in Table 10, a SCNN-S model using SIG is better than a SCNN-S model using REL. Thus, a SCNN model with Inception-v4 using the commonly used REL is not always the best. 

\section{Conclusions}

A MCNN with a more variety of different activation functions can achieve better performance than a SCNN. Simulation results show that all of the tested MCNN-SM, MCNN-MS, MCNN-MM, and MCNN-M models performed better than all of the tested SCNN models (except for one case which was a tie) in terms of image classification test accuracy. Interestingly, many MCNN models and even some SCNN models using an activation function other than ReLU were able to outperform SCNN models using ReLU. Therefore, using the same activation function for all AL neurons and all FC hidden neurons may not always be optimal for CNNs. Since there are many more candidate MCNN models than SCNN models, MCNN models have much more chances of performing better with an optimized set of activation function combinations than SCNN models. Although the differences in the average training and test accuracies for MCNN and SCNN models may not seem significant right now, MCNN models can at least be considered as an alternative variant of traditional CNNs to achieve better performance. Even the smallest difference may eventually become significant. The test accuracy improvement for MCNN models and the small numbers of MCNN models tested show that it is feasible and not difficult to find MCNN models that perform better than SCNN models. MCNN models have the ability to perform much better than SCNN models especially for various complex applications with big complex image data.

\section{Future works}

\subsection{MCNN model optimization}

More MCNN models based on currently powerful architectures such as ResNets, DenseNets, and DPNs will be created and tested. An automatic process for building, training, testing, and optimizing SCNN and MCNN models will be created. Different neurons on each AL and different neurons on each FC hidden layer may also use different activation functions. Then, there are a total of \textit{$j\textsuperscript{m}k\textsuperscript{n}$-jk} different MCNN models based on Table 1. Thus, it is not practical to evaluate all models to find the best model, so a partial number of models can be created by randomly choosing activation functions for all neurons and then evaluated. A general method will be developed and then used to make a complex multi-function Inception-v4 architecture by assigning an activation function from a set of activation functions to each individual AL in each CB.

Better and faster algorithms will be developed to efficiently find the best set of different activation functions for all neurons of a MCNN to achieve the best performance for various important applications such as medical imaging for cancer detection [4] and brain imaging for mental illness diagnosis such as autism detection [25]. Developing intelligent high-speed optimization software for identifying the best MCNN model for a particular application will be a difficult long-term challenge.

\subsection{CNN software enhancement}

Current CNN software libraries such as Keras and scikit-learn can be used to create MCNN models by choosing different activation functions for ALs and FC hidden layers. However, they have the following limitation: an AL or FC hidden layer can only use one activation function for all neurons. Then, these libraries can be improved by allowing different neurons in an AL or FC hidden layer to use different activation functions.

Existing CNN software systems based on the single-function deep convolutional architecture such as Inception-v4 and Inception-ResNet-v2 [18] can be modified by choosing different activation functions for different neurons in all ALs and FC hidden layers to create MCNN software systems. Since the MCNN model selection among a large number of candidate MCNNs would take a very long computational time for big image data, parallel algorithms will be developed to speed up the large-scale optimization process.

\subsubsection*{Acknowledgments}

Data were provided in part by OASIS: Cross-Sectional: Principal Investigators: D. Marcus, R, Buckner, J, Csernansky J. Morris; P50 AG05681, P01 AG03991, P01 AG026276, R01 AG021910, P20 MH071616, U24 RR021382.

\section*{References}

\small

[1] LeCun, Y., Bengio, Y.\ \& Hinton, G.E. (2015) Deep learning. Nature 521, pp.\ 436--444.

[2] Krizhevsky, A., Sutskever, I.\ \& Hinton, G.E.\ (2012) Imagenet classification with deep convolutional neural networks. In {\it Advances in Neural Information Processing Systems 25}, pp.\ 1097--1105. Cambridge, MA: MIT Press.

[3] He, K., Zhang, X., Ren, S.\ \& Sun, J. \ (2016) Deep Residual Learning for Image Recognition. In Proceedings of the 2016 IEEE Conference on Computer Vision and Pattern Recognition (CVPR), pp.\ 770--778.

[4] Esteva, A., Kuprel, B., Novoa, R.A., Ko, J., Swetter, S.M., Blau, H.M.\ \& Thrun, S.\ (2017) Dermatologist-level classification of skin cancer with deep neural networks. {\it Nature} {\bf 542}(7639):115--118.

[5] Nugraha, B.T, Su, S.-F.\ \& Fahmizal, F.\ (2017) Towards self-driving car using convolutional neural network and road lane detector. In Proceedings of the 2nd International Conference on Automation, Cognitive Science, Optics, Micro Electro-Mechanical System, and Information Technology (ICACOMIT), pp.\ 65--69. 

[6] Silver, D., Huang, A., Maddison, C.J., Guez, A., Sifre, L., Driessche, G.V.D., Schrittwieser, J., Antonoglou, I., Panneershelvam, V., Lanctot, M., Dieleman, S., Grewe, D., Nham, J., Kalchbrenner, N., Sutskever, I., Lillicrap, T., Leach, M., Kavukcuoglu, K., Graepel, T.\ \& Hassabis, D.\ (2016) Mastering the game of Go with deep neural networks and tree search. Nature (529), pp 484--503. 

[7] Fukushima, K.\ (1979) Neural network model for a mechanism of pattern recognition
unaffected by shift in position-Neocognitron. {\it Transactions of the IECE} {\bf J62-A}(10):658--665.
 
[8] LeCun, Y., Bottou, L., Bengio, Y.\ \& Haffner, P.\ (1998) Gradient-based learning applied to document recognition. {\it Proc. IEEE} {\bf 86}(11):2278--2324.

[9] {\it Large Scale Visual Recognition Challenge 2012 (ILSVRC2012)} (2012)  [Online.] Available: http://www.image-net.org/challenges/LSVRC/2012/results.html.

[10] {\it Large Scale Visual Recognition Challenge 2014 (ILSVRC2014)} (2014)  [Online.] Available: http://www.image-net.org/challenges/LSVRC/2014/.

[11] {\it Large Scale Visual Recognition Challenge 2015 (ILSVRC2015)} (2015) [Online.] Available: http://www.image-net.org/challenges/LSVRC/2015/index.

[12] {\it Large Scale Visual Recognition Challenge 2017 (ILSVRC2017)} (2017) [Online.] Available: http://www.image-net.org/challenges/LSVRC/2017/index.

[13] {\it COCO 2015 Object Detection Task} (2015) [Online.] Available: http://cocodataset.org/\#detection-2015.

[14] He, K.\ (2016) 
Deep Residual Networks - 
Deep Learning Gets Way Deeper. 
[Online.] Available: 
https://icml.cc/2016/tutorials/icml2016\textunderscore tutorial\textunderscore deep\textunderscore residual\textunderscore networks\textunderscore kaiminghe.pdf.

[15] Szegedy, C., Liu, W., Jia, Y., Sermanet, P., Reed S., Anguelov D., Erhan, D., Vanhoucke, V.\ \& Rabinovich, A.\ (2015) Going Deeper with Convolutions. In Proceedings of 2015 IEEE Conference on Computer Vision and Pattern Recognition (CVPR), pp.\ 1--9. 

[16] Huang, G., Liu Z., Maaten, L.v.d.\ \& Weinberger, K.Q. (2018) Densely Connected Convolutional Networks. [Online]. Available:  https://arxiv.org/abs/1608.06993.

[17] Chen, Y., Li, J., Xiao, H., Jin, X., Yan, S.\ \& Feng J. (2018) Dual Path Networks. [Online]. Available:  https://arxiv.org/abs/1707.01629. 

[18] Szegedy, C., Ioffe, S., Vanhoucke, V.\ \& Alemi, A.\ (2017) Inception-v4, Inception-ResNet and the Impact of Residual Connections on Learning. In Proceedings of the Thirty-First AAAI Conference on Artificial Intelligence (AAAI-17), pp.\ 4278--4284.

[19] Clevert, D.-A., Unterthiner, T.\ \& Hochreiter, S. (2016) FAST AND ACCURATE DEEP NETWORK LEARNING BY EXPONENTIAL LINEAR UNITS (ELUS). [Online]. Available:  https://arxiv.org/abs/1511.07289.

[20] Zhang, L.M. (2016) A new multifunctional neural network with high performance and low energy consumption. 15th IEEE International Conference on Cognitive Informatics and Cognitive Computing, pp.\ 101--109.

[21] LeCun, Y., Cortes, C.\ \& Burges, C. (n.d.). \textit{MNIST handwritten digit database}. [Online]. Available: http://yann.lecun.com/exdb/mnist/.

[22] {\it Digit recognizer} (n.d.). [Online]. Available: https://www.kaggle.com/c/digit-recognizer.

[23] {\it OASIS Brains Datasets}. [Online]. Available:  http://www.oasis-brains.org/\#data.

[24] Marcus, D.S., Wang, T.H., Parker, J., Csernansky, J.G., Morris,
J.C.\ \& Buckner, R.L.\ (2007) Open access series of imaging studies (oasis):
cross-sectional MRI data in young, middle aged, nondemented, and
demented older adults. Journal of cognitive neuroscience 19(9),
1498--1507.

[25] Hazlett, H.C., Gu, H., Munsell B.C., Kim, S.H., Styner M., Wolff, J.J., Elison, J.T., Swanson, M.R., Zhu, H., Botteron, K.N., Collins, D.L., Constantino, J.N., Dager, S.R., Estes, A.M., Evans, A.C., Fonov, V.S., Gerig, G., Kostopoulos, P., McKinstry, R.C., Pandey, J., Paterson, S., Pruett, J.R., Schultz, R.T., Shaw, D.W., Zwaigenbaum, L., Piven J.\ \& The IBIS Network (2017) Early brain development in infants at high risk for autism spectrum disorder. Nature 542, pp.\ 348--351.

 \end{document}